\pdfoutput=1

\documentclass[11pt]{article}

\usepackage[dvipsnames]{xcolor}

\usepackage{acl}

\usepackage{times}
\usepackage{latexsym}
\usepackage{comment}
\usepackage{tabularx}
\usepackage{subcaption}

\usepackage{graphicx}
\usepackage{multicol}
\usepackage{multirow}
\usepackage{color,soul}

\usepackage[T1]{fontenc}

\usepackage[utf8]{inputenc}

\usepackage{microtype}

\usepackage{inconsolata}

\title{Towards Detecting Harmful Agendas in News Articles\\
{\small \textcolor{orange}{Warning: This paper contains examples of online text containing strong views and offensive language.}}}

\author{Melanie Subbiah$^*$ \\
  Columbia University \\
  \texttt{m.subbiah@columbia.edu}\\\And
  Amrita Bhattacharjee$^*$ \\
    Arizona State University \\
  \texttt{abhatt43@asu.edu} \\\And
    Yilun Hua \\
  Columbia University \\
  \texttt{yh3228@columbia.edu} \\\AND
Tharindu Kumarage \\
  Arizona State University \\
  \texttt{kskumara@asu.edu} \\\And
  Huan Liu \\
  Arizona State University \\
  \texttt{huanliu@asu.edu} \\\And
    Kathleen McKeown \\
  Columbia University \\
  \texttt{kathy@cs.columbia.edu}
  }

\begin{document}
\maketitle
\def\thefootnote{*}\footnotetext{These authors contributed equally to this work.}\def\thefootnote{\arabic{footnote}}
\begin{abstract}
Manipulated news online is a growing problem which necessitates the use of automated systems to curtail its spread. We argue that while misinformation and disinformation detection have been studied, there has been a lack of investment in the important open challenge of detecting harmful agendas in news articles; identifying harmful agendas is critical to flag news campaigns with the greatest potential for real world harm. Moreover, due to real concerns around censorship, harmful agenda detectors must be interpretable to be effective. In this work, we propose this new task and release a
 dataset, \textsc{NewsAgendas}, of annotated news articles for agenda identification. We show how interpretable systems can be effective on this task and demonstrate that they can perform comparably to black-box models.
\end{abstract}

\section{Introduction}
In recent years, the spread of misinformation and disinformation has become a particularly persistent and harmful issue online \cite{bastick2021would, mueller2020internet, vosoughi2018spread, zhang2020overview}. For example, during the COVID-19 pandemic in the United States, we saw several instances of malicious actors propagating disinformation regarding mask mandates, vaccines, and fake remedies and cures to discredit the government and public health officials. The people initiating these disinformation campaigns typically have some harmful agenda, such as discrediting an individual/group or encouraging disruptive real-world action. Furthermore, with new conversational language models such as ChatGPT and GPT-4 \cite{openai2023gpt4}, a malicious actor can generate human-like harmful text content at scale.

\begin{table}[ht]
\centering
\begin{tabularx}{.48\textwidth}{p{7.5cm}}
    \hline
    \small \textbf{Title: } Are You Prepared For The War To End All Wars? \\ \small ... \colorbox{BurntOrange}{Everyone in the know (global elitists) knows what is} \colorbox{BurntOrange}{happening, just not exactly when they can pull it all} \colorbox{BurntOrange}{together}. And the media awaits this war with baited breath as they count down to the dramatic moment when they can report the incident that will compel the innocent to attack the guilty. Anyone with half a brain can see the greatly increased anti-Russian propaganda of the past few weeks. \colorbox{GreenYellow}{This has} \colorbox{GreenYellow}{happened as the Russia-gate claims have fallen to pieces,} \colorbox{GreenYellow}{as former CIA analyst Raymond McGovern, the late} \colorbox{GreenYellow}{Robert Parry, Paul Craig Roberts, and others have} \colorbox{GreenYellow}{documented so assiduously.}\colorbox{SkyBlue}{All across the media} \colorbox{SkyBlue}{spectrum, from the big name corporate stenographers} \colorbox{SkyBlue}{like The New York Times, CNN, National Public Radio,} \colorbox{SkyBlue}{The Washington Post to The Atlantic and Nation} \colorbox{SkyBlue}{magazines and other leftist publications such as Mother} \colorbox{SkyBlue}{Jones and Who What Why, the Russia and Putin bashing} \colorbox{SkyBlue}{has become hysterical in tone ...}
\\\hline
\end{tabularx}
\caption{Example article with annotated spans from our dataset, original article from \textit{infiniteunknown.net}, a source with label \textit{conspiracy} in the FakeNewsCorpus. Orange spans are annotated as \textit{conspiracy}, yellow spans are \textit{political bias}, and blue spans are \textit{propaganda}.}
\label{tab:articleexamples}
\end{table}

Identifying these types of harmful news campaigns typically requires consideration of three important attributes:
\begin{enumerate}
    \itemsep0em 
    \item \textbf{Factuality} - Does the article rely on false information?
    \item \textbf{Authorial Deception} - Did the author knowingly deceive the reader?
    \item \textbf{Agenda} - Why did the author deceive the reader?
\end{enumerate}
Misinformation in news is any article which relies on false information and can therefore be identified by focusing on \textit{factuality}. Disinformation is deliberately misleading information created/disseminated with an intent to deceive \cite{shu2020combating}, so can be identified by \textit{factuality} and \textit{authorial deception}. However, the degree of harm caused by disinformation and misinformation depends on the \textit{agenda} (or goal) of the article. \citet{fallis2015disinformation} advocates for this kind of focus on agenda as a useful marker of intentionality in disinformation detection. Defining what constitutes a harmful agenda is an inherently subjective task and requires a notion of good and bad. Researchers in different domains have tried defining and formalizing the concept of harm, such as harmful online content \cite{scheuerman2021framework}, COVID-19 related tweets~\cite{alam2021fighting}, etc. However, to the best of our knowledge, the notion of harmful agendas in journalistic news articles has not been explored yet. In this paper, we therefore propose a new task of detecting harmful agendas in news articles. Inspired by definitions of harm in other works, we specifically focus on real-world harm, meaning articles that spur core belief change or actions that significantly harm someone. 

To develop an initial detector, we formulate this task as classifying an article's agenda as \textit{harmful} or \textit{benign}, based solely on the article text, and we annotate a dataset, \textsc{NewsAgendas}, to evaluate performance. We note that future work could also formulate this problem in several other ways, such as also identifying the target audience, or additionally using metadata or contextual cues such as author information, publication platform, etc.

We imagine this type of agenda detector could be used to flag potentially harmful articles for further inspection. It is therefore critical that any such detector be interpretable so that further examination could quickly reveal why an article was flagged and screen out any falsely identified articles. For sensitive application areas, there is a need to build models that are interpretable by design, rather than trying to interpret their results after the fact \cite{rudin2019stop}. Given the subjectivity and the sensitivity of this task, we build an interpretable model that uses 
extractive rationale-based feature detectors to ensure faithfulness and interpretability, not only at the feature level but also at the text level. 

Our primary contributions are:
\begin{enumerate}
    \itemsep0em 
    \item Defining the important open challenge of detecting harmful agendas in news articles.
    \item Annotating and releasing \textsc{NewsAgendas} - 506 news articles, encompassing 882 fine-grained label annotations for this task.\footnote{All data and code is available at \href{https://github.com/melaniesubbiah/harmfulagendasnews}{https://github.com/melaniesubbiah/harmfulagendasnews}.}
    \item Developing a harmful agenda detector which jointly prioritizes interpretability and performance.
\end{enumerate}

\section{\textsc{NewsAgendas} Dataset}
\label{ssec:eval-data}
\begin{table*}[hbt!]
    \centering
    \begin{tabularx}{.98\textwidth}{p{1.62cm}|p{5.3cm}|p{7.5cm}}
    \hline
         \textbf{ Label} & \textbf{ Definition} & \textbf{ Notes on Connection to Article Agenda} \\\hline\hline
          Clickbait & \small An exaggeration or twisting of information to shock and grab the attention of the reader. & \small Can be used to promote a harmful agenda (\citeauthor{health_feedback_2021, chen2015misleading}), but often just a marketing strategy which is relatively benign.\\\hline
         Junk \,\,\,\,Science & \small Untested or unproven theories presented as scientific fact. & \small Can be unintentional, but has a high potential for harm, particularly in the medical domain (\citeauthor{junksci_1, poynter_1}).\\\hline
          Hate Speech & \small Language that promotes or justifies hatred, violence, discrimination, or negative prejudice against a person or category of people. & \small Involves extreme language that indicates clear intent on the part of the author and has a high potential for harm, even physical violence (\citeauthor{hatespeech_1}).\\\hline
         Conspiracy Theory &  \small A belief that some covert but influential organization is responsible for a circumstance or event. & \small Erodes public trust in science, institutions, and government (\citeauthor{ahmed2020four, oliver2014medical}) which may not be intentional on the part of individual actors but is harmful.\\\hline
         Propaganda & \small Promoting or publicizing a particular political cause or perspective. & \small Polarizes readers and harms the democratic environment necessary for healthy political debate (\citeauthor{guarino2020characterizing}).\\\hline
         Satire & \small Using humor, irony, or exaggeration to critique something or to amuse. & \small Not typically harmful when used to reveal a social/political truth, rather than for hate (\citeauthor{levi2019identifying, golbeck2018fake}).\\\hline
         Negative Sentiment & \small Evokes a negative emotional response in the reader. & \small Evoking negative emotionality can create a lasting reaction (\citeauthor{emotion}), which can be more benign like sensationalism (\citeauthor{emotion2}), or more harmful like negative propaganda.\\\hline
         Neutral Sentiment & \small Generally neutral/factual tone throughout the article. Does not evoke strong emotion. & \small Credible news organizations often have guidelines for objective and neutral reporting of `hard-news' (\citeauthor{obj_1}). \\\hline
         Positive Sentiment & \small Evokes a positive emotional response in the reader. & \small Research suggests positive sentiment is not often used in disinformation or to instigate/polarize readers (\citeauthor{alonso2021sentiment}).\\\hline
         Political Bias &  \small Angling information toward a particular political cause or perspective. & \small Biased articles may misrepresent/slant facts to support (harmful) agendas in cases of contentious topics (\citeauthor{chen2020detecting}).\\\hline
          Call to Action & \small Urging the reader to do (or not do) something in order to further some goal. & \small Instigating or urging the reader to take some action for example via bandwagoning (\citeauthor{da2019fine}) may result in a (harmful) real-world effect. \\\hline
    \end{tabularx}
    \caption{The definitions for the full set of labels annotators were asked to label articles with.}
    \label{tab:question2}
\end{table*}

In order to evaluate our model's performance and contribute an initial benchmark for this task, we annotated news articles which we are releasing as a novel dataset, \textsc{NewsAgendas}. 

\subsection{Features of Interest}

To promote interpretability, we hypothesize based on consultation with journalism professors at Arizona State University 
that the features shown in Table \ref{tab:question2} (e.g., hate speech, propaganda, etc.) may have a significant relationship to the overall classification of article agenda in the sociopolitical context of the United States (see Table \ref{tab:question2} for justification).

We are therefore interested in annotating these feature labels at the article-level as well as the overall agenda classification for the article. Using these features also allows us to build on the training datasets used in fine-grained news classification to classify news into these different categories.

\subsection{Articles}
\label{articles}
We use articles from the FakeNewsCorpus\footnote{\href{https://github.com/several27/FakeNewsCorpus}{https://github.com/several27/FakeNewsCorpus}} along with satire and real news articles from the \citet{yang_satirical_2017} dataset and propaganda articles from the Proppy corpus \cite{barron2019proppy} to cover a range of articles that should contain the features and agendas we are interested in. The FakeNewsCorpus contains articles in English from a web scrape of sources which frequently post misinformation. Each source has one or more specific labels indicating the general type of content it publishes and many of these labels match 
our features of interest (e.g., junk science, conspiracy theories, etc.). Since these labels are assigned at the source level, they serve as weak labels at the article level. 
We sample 600 articles for annotation, sampling to match the distribution of weak labels in the FakeNewsCorpus (based on the articles' primary weak labels; see Appendix E for more detail).

\subsection{Annotation Method}
We hired Columbia University students who study journalism, political science, or natural language processing 
and thus have experience interpreting news (see Appendix B for hiring details). 

We presented each annotator with the title of the article and the first 1,700 characters of the article truncated to the last sentence. They were asked to assume the article contained some false claims, and then rate whether it advanced a harmful agenda on a scale of 1 to 5. We allowed for some subjective interpretation of what a \textit{harmful agenda} meant, but we prompted them to think of the scale of impact and whether an article might promote a real-world negative action or a strong negative belief about an individual or group of people. Lastly, they were asked to label the features found in Table \ref{tab:question2}, with the associated definitions provided, and provide 1-3 supporting evidence spans from the article for each label. They were prompted to first consider the article's primary weak label, and not to exhaustively label features. Since the features and score were labeled separately, we did not enforce any particular relationship between an individual feature and the overall label. See Appendix C for the full task instructions. We asked them to annotate a broader list of features than we used in our models for this paper to enable future work on this problem.

The full evaluation dataset, \textsc{NewsAgendas}, consists of 506 annotated articles with 882 fine-grained label annotations. 
Each article additionally has its original weak label. See Appendix D for the label and score distribution and dataset examples. 

\subsection{Annotation Quality}
To measure agreement between annotators, we held out an additional 90 articles for annotation by at least 2 graduate students (on average 3.4 students per article) studying 
natural language processing or journalism. We asked annotators just to label the harmful agenda score and to identify whether a specific feature from Table \ref{tab:question2} was present. For each feature, we presented 5 articles with that weak label and 5 random articles. For sentiment, we presented this task as a 3-way classification between positive, neutral, and negative (see Appendix C for full task instructions). We then computed 
Cronbach's alpha (a measure of internal consistency \cite{cronbach1951coefficient}) across the annotators' responses. We observed good agreement across the harmful agenda scores (Table \ref{tab:annot}), and moderate agreement across the individual feature labels. These results indicate the data is of reasonable quality but future work could place more emphasis on how to well-annotate some of the trickier features.

\begin{table}[ht]
\centering
\begin{tabular}{l|c}
    \hline Annotation Type  & Cronbach's Alpha  \\\hline\hline
     Harmful Agenda Scores &0.78 (0.69, 0.84)\\\hline
     Feature Labels&0.53 (0.35, 0.67)\\\hline
\end{tabular}
\caption{Cronbach's alpha consistency measure for the annotated scores and feature labels in the annotation quality experiments. 95\% confidence intervals are shown in parentheses. As a reference, randomly generated scores/labels produce a Cronbach's Alpha <0.06.}
\label{tab:annot}
\end{table}

\subsection{Labels}
We define different sets of feature labels used in the paper for clarity:
\begin{enumerate}
    \item \textbf{Annotated gold labels} - Feature labels assigned by our annotators in \textsc{NewsAgendas}. 
    \item \textbf{Weak labels} - Feature labels assigned at the source-level from the FakeNewsCorpus.
    \item \textbf{BERT/FRESH labels} - Feature labels predicted by our trained models (seen in Sec.~\ref{results}).
\end{enumerate}
The annotated gold labels are the standard which we can evaluate our system against, but we cannot train on them since there is not enough data per label and we cannot contaminate evaluation results by training on the evaluation data. We therefore use the weak labels for training, since there is a large quantity of weak labelled articles, although they are not as accurate.

\section{Methods}
We leverage large weakly labeled datasets to train feature classifiers for our features of interest. We prioritize exploring different levels of interpretability in the models we compare and what performance tradeoffs come at each level. To focus our analysis, we select \textbf{7 features} to study in-depth: \textit{clickbait}, \textit{junk science}, \textit{hate speech}, \textit{conspiracy theories}, \textit{propaganda}, \textit{satire}, and \textit{negative sentiment}. Out of the 4 features we excluded, 3 did not have enough labelled data. For the 4th, political bias, after consulting our journalism experts, we determined the relationship between harmful agendas in news articles and political bias is nuanced and needs further study. We therefore leave political bias to future work to promote simplicity and interpretability in our approach.

\subsection{Models}

\begin{figure*}
\includegraphics[width=\textwidth]{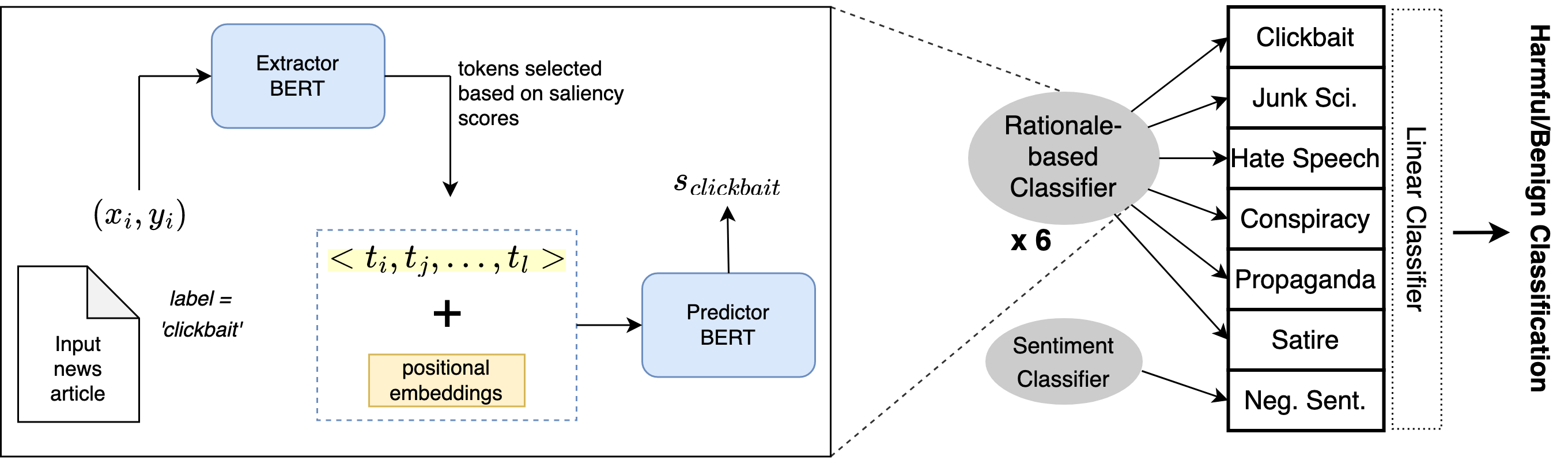}
\caption{The full system consists of 6 different rationale-based classifiers and a simple sentiment classifier. These 7 features are input to a linear classifier which outputs the final classification.} 
\label{fig:model}
\end{figure*}

As shown in Figure \ref{fig:model}, our approach is to separately train individual neural feature classifiers for each of the 7 features of interest. We then combine these features using a linear classifier to produce the final agenda classification. Our model is interpretable at the final level since the feature vector indicates the features that contribute to the final classification. It is also interpretable at the feature-level, where 6 of 7 features are derived from rationale-based models, which indicate the subset of input tokens that contribute to the feature classification.

Since we want to ensure faithfulness and interpretability, we derive our rationale model from the FRESH framework \cite{jain_learning_2020} (see Figure \ref{fig:model}). We first finetune a BERT model (\texttt{Extractor BERT}) to predict a feature label from the full article text. For each token in the document, we derive a saliency score from the [CLS] token attention weights in the penultimate layer of this extractor. We extract as a rationale the top 20\% most important tokens (with respect to saliency scores), irrespective of contiguity (each word is treated independently). Next, we finetune a second BERT model (\texttt{Predictor BERT}) to predict the feature label using only these extracted rationale tokens concatenated as input. This approach differs from the original FRESH paper in that we do not use a human-annotated dataset to introduce additional token-level supervision in rationale extraction. We also modify the FRESH framework to leverage positional embeddings for tokens. See Appendix A for details on training hyperparameters.

For the sentiment classification, we use the VADER classifier built into the NLTK Python library \cite{hutto2014vader, loper2002nltk}. We choose VADER over more recent LLM-based sentiment analysis models, to facilitate interpretability. We compute the compound polarity score on a concatenation of the article title and contents. Articles with a compound score less than 0 are labeled as negative.

\subsection{Training Data}
For training data for the individual feature detectors, we use articles and weak labels from the same datasets described in Section \ref{articles} (however, we remove any articles used in \textsc{NewsAgendas}).  
We handle negative sentiment labels at the model level (discussed in the next section).

Since the FakeNewsCorpus was collected from a broad scrape of unreliable websites, we noticed many of the texts did not fit the format of a news article. We therefore only use articles from the FakeNewsCorpus whose source overlaps with the list of sources used by NELA-GT \cite{gruppi2021nela} or \citet{li2020mm}'s Covid-19 dataset in order to filter for high quality sources.  While this approach is not exhaustive, it significantly improves the quality of the data since the sources are validated by multiple misinformation datasets. We also search and remove URLs and variants of the source names from the articles to avoid model memorization of source-label pairings.

For each individual feature detector's training dataset, we sample 2,500 articles with the feature label we hope to detect (positive examples), and sample a range of negative examples based on a set of criteria (see Appendix E for details on negative examples for each feature). 
For each label, we adopt a weighted sampling strategy to increase the diversity of sources. We assign each article from a website $w$ a weight $\frac{1}{c_w}$, where $c_w$ is the total count of articles from website $w$. We then normalize these weights to sum to 1.

We additionally hold out 500 articles for the dev set and 500 articles for the test set. The test set articles come from a different set of websites than were used for the train and dev sets to make sure the test scores can not be inflated by any model memorization of website-specific styles.

\section{Results}
\label{results}
We investigate a series of research questions that analyze the efficacy of our overall approach, as well as individual components in our dataset and models.

\subsection{How well can we predict the overall agenda score?}
\label{ssec:pred-mal}
We experiment with predicting the \textsc{NewsAgendas} annotated agenda score using different variants of our system. We fit the final logistic regression layer to the data using 10-fold cross-validation. The input is the 7 binary feature labels and the output is a binary classification of harmful or benign agenda - we bucket agenda scores 1-3 as benign and 4-5 as harmful (annotators gave a score of 3 when they were unsure of whether there was a harmful agenda in the text). We compare our method using the predicted features against three baselines: (1) predicting the majority class (0-benign), (2) using the weak source-level feature labels for logistic regression, and (3) finetuning a BERT model to classify the agenda (see Table \ref{tab:classification}). Baseline (2) demonstrates how this approach may be limited by the quality of the weak labels. Baseline (3) demonstrates a comparison against a fully black box model. We additionally compare against logistic regression using the annotated gold labels as an oracle. Using the annotated gold labels indicates a rough upper bound on performance for this type of feature-based approach, but could not be used in practice since it relies on a human annotating the articles. Note that the performance of the oracle implies a significant scope for improvement, and re-affirms our hypothesis that detecting harmful agendas in news articles is an especially difficult task for an automated system. 

\begin{table}[ht]
\centering
\begin{tabular}{l|c|c}
    \hline Method &Accuracy & Bal. Acc.  \\\hline\hline
     Oracle Logistic Reg. & 76.7\,\,\,\,\,\,\,\,\,\,\,& 75.6\,\,\,\,\,\,\,\,\,\,\,\\\hline\hline
     Predict Majority Class &58.4\,\,\,\,\,\,\,\,\,\,\, & 50.0\,\,\,\,\,\,\,\,\,\,\, \\\hline
     Weak Logistic Reg. &58.9\,\,\,\,\,\,\,\,\,\,\, & 58.4\,\,\,\,\,\,\,\,\,\,\,\\\hline
     BERT-based Baseline&63.0 \small(0.21)&62.2 \small(0.36)\\\hline
     BERT System (Ours) &60.1 \small(0.02) & 60.1 \small(0.02)\\\hline
     FRESH System (Ours) &59.3 \small(0.03) & 59.3 \small(0.03)\\\hline

\end{tabular}
\caption{
Overall performance (accuracy and balanced accuracy) on detecting harmful agendas in \textsc{NewsAgendas}. The oracle logistic regression uses the annotated gold labels. Results are averaged with standard deviation shown in parentheses for neural models.}
\label{tab:classification}
\end{table}
    
 The oracle logistic regression model with the human annotated gold labels performs well, indicating our features of interest are very useful for the ultimate classification and promote interpretable classification of article agenda. 
 The three systems we compare (with three different levels of interpretability) all perform better than both the majority baseline and logistic regression using just article weak labels.
 We also see that while we lose a little performance for every increase in interpretability (differences shown in table are statistically significant by a two-sample t-test, p<.0001), 
 it is possible to build interpretable models that are almost as effective as the black box models for this task. 
 This interpretability is critical because a real-world system with this accuracy would require human oversight. The strong results of the oracle model also demonstrate that investing in better feature detectors could result in improved overall agenda classification, even beyond the black-box approach.

\subsection{How are the features in \textsc{NewsAgendas} related to the overall agenda score?}



We first perform a pairwise analysis of which labels are more related to higher agenda scores over others in \textsc{NewsAgendas}, using a pairwise Wilcoxon test. 
\textit{Hate speech} and \textit{negative sentiment} are associated with higher scores most often over other labels, suggesting that these two features are particularly strong indicators of a harmful agenda. Interestingly, \textit{call to action} loses this pairwise comparison most often, even though it seems this label would be the biggest indicator of the article encouraging a real-world outcome. This may be because \textit{call to action} was the least represented feature in the data (only labeled 8 times) so there is not a lot of data on this feature. \textit{Neutral sentiment} and \textit{satire} are associated with lower scores most often over other labels, suggesting that these two features are stronger indicators of a benign agenda.
See Appendix F for more details on this analysis.

\begin{table}[ht]
\centering
\begin{tabular}{l|c|c|c|c}
    \hline Feature&Annot.&Weak&BERT&FRESH\\\hline\hline
     Clickbait&0.96&0.12&0.47&0.45\\\hline
     Junk Sci.&0.22&-0.16&-0.28&-0.45\\\hline
     Hate Sp.&\textbf{1.76}&0.21&\textbf{0.57}&\textbf{0.61}\\\hline
     Conspir.&0.86&0.20&-0.24&-0.07\\\hline
     Propagan.&1.31&\textbf{0.80}&0.55&0.42\\\hline
     Satire&0.62&0.34&0.22&0.17\\\hline
     Negative&1.55&n/a&0.53&0.52\\\hline
\end{tabular}
\caption{Weights for each feature learned by the logistic regression models across different feature label sets. The weights are averaged across the different cross-validation subsets and across seeds when appropriate.}
\label{tab:ablations}
\end{table}

We also look at the weights learned by the final logistic regression layer over the features to determine what relationship the models learn between the feature labels and the final harmful agenda score. We see that almost all of the models place the highest weight (noted in bold) on \textit{hate speech} with \textit{negative sentiment} and \textit{propaganda} generally coming in second. The models generally place the lowest weights on \textit{junk science}, \textit{conspiracy theories}, and \textit{satire}.

\subsection{How well do our feature detectors work?}
In order to evaluate how well each feature classifier learned its training task (predicting the weak label from the FakeNewsCorpus for its feature), we evaluate predicted labels against weak labels across three datasets:
1) the validation set, 2) the test set, and 3) \textsc{NewsAgendas}. We compare the FRESH-based models relative to the baseline of just using the fine-tuned extractor BERT model to predict the label to explore different levels of interpretability.

In Table \ref{tab:predictors}, we see that the feature classifiers generalize effectively to articles from new sources in the test set, although the performance drop (relative to the validation set) indicates that the models are relying on some source-specific qualities of articles during training. We also see reasonable performance on the articles in \textsc{NewsAgendas} with the exception of the satire model which performs poorly. We think the poor satire performance is because the training satire articles came from higher quality websites than many of the sites in the FakeNewsCorpus and therefore the text style may be too different to transfer to many of the articles in \textsc{NewsAgendas}.

\begin{table*}[h]
    \centering
    \begin{tabular}{l||c|c||c|c||c|c}
    & \multicolumn{2}{c||}{Val. Set} 
    & \multicolumn{2}{c||}{Test Set} & \multicolumn{2}{c}{\textsc{NewsAgendas}} \\\hline
    Feature & BERT & FRESH & BERT & FRESH & BERT & FRESH \\
    \hline \hline
    Clickbait &\textbf{90.5} \small(0.8)&88.7 \small(0.6)& \textbf{61.1} \small(1.3)& 59.0 \small(0.3)&\textbf{76.9} \small(0.5)& 71.6 \small(3.6)    \\
    \hline
    Junk Science &\textbf{93.9} \small(0.7) &93.0 \small(0.8) & 89.3 \small(0.8) & \textbf{89.5} \small(0.7)& \textbf{77.4} \small(1.6)& 73.8 \small(2.5)  \\
    \hline
    Hate Speech &\textbf{91.7} \small(0.3) &90.8 \small(0.8) & 83.0 \small(1.3)& \textbf{83.4} \small(0.8)&\textbf{65.4} \small(0.7)& 64.4 \small(1.1)  \\
    \hline
    Conspiracy Theory &\textbf{94.2} \small(0.3) &93.2 \small(0.6)& \textbf{74.9} \small(1.0)& 74.3 \small(1.6)& \textbf{62.7} \small(1.5) & 61.7 \small(1.4)  \\
    \hline
    Propaganda & \textbf{91.9} \small(0.6)& 91.1 \small(0.3) & 70.4 \small(1.1) & \textbf{71.5} \small(2.2)&\textbf{77.3} \small(1.3)& 73.1 \small(1.8)\\
    \hline
    Satire & \textbf{95.9} \small(0.2) & 94.5 \small(0.6)& 66.9 \small(2.1)& \textbf{73.1} \small(2.4)&\textbf{51.5} \small(0.4) & \textbf{51.5} \small(0.6)  \\
    \hline
    \end{tabular}
    \caption{Mean balanced accuracy scores (standard deviation in parentheses) for predicting the weak labels using the BERT and FRESH feature classifiers.}
    \label{tab:predictors}
\end{table*} 

We then evaluate how well the predicted labels agree with the annotated gold labels. To measure overlap between predicted labels and annotated gold labels, we report the intersection-over-union (IOU) and the recall for the classifiers (see Table \ref{tab:agreement}). As a baseline, we include the agreement between the weak labels and the annotated gold labels. The generally low weak label agreement shows that the source-level labels for articles provide fairly distant supervision relative to human judgment. We see that the BERT and FRESH models have worse but fairly similar overlap as the weak labels in many cases. The junk science and satire models have the least overlap. The black-box BERT model seems to have a slight advantage on the FRESH model, indicating there is an interpretability/performance tradeoff. 

\begin{table*}[hbt!]
\centering
    \begin{tabular}{l||c|c|c||c|c|c}
    & \multicolumn{3}{c||}{IOU} & \multicolumn{3}{c}{Recall-1} \\\hline
    Feature & Weak & BERT & FRESH & Weak & BERT & FRESH \\
    \hline \hline
    Clickbait & \textbf{32.0} &30.9 \small(0.7) &25.3 \small(1.6) & \textbf{53.3} & 46.4 \small(1.3)& 40.2 \small(8.5)\\\hline
    Junk Science & \textbf{18.5} & 17.1 \small(0.8) &12.5 \small(4.2)& 41.7 & 75.0 \small(6.8)&\textbf{77.8} \small(7.8)\\\hline
    Hate Speech & 16.5 &18.0 \small(1.3)&\textbf{19.1} \small(2.8)&34.7 & \textbf{64.6} \small(2.5)&57.1 \small(7.2)\\\hline
    Conspiracy & \textbf{27.7}&18.8 \small(0.5)&18.4 \small(0.3)&  40.3& 60.1 \small(1.2)&\textbf{67.5} \small(7.4)\\\hline
    Propaganda & \textbf{56.2}  &43.0 \small(1.6)&40.1 \small(2.7) & \textbf{77.1} & 60.0 \small(2.5)&59.8 \small(4.5)\\\hline
    Satire & \textbf{47.9}  &\,\,\,2.8 \small(0.8) &\,\,\,2.3 \small(0.8) & \textbf{61.4} &\,\,\,2.9 \small(2.4)&\,\,\,2.4 \small(0.8)\\\hline
    Negative Sentiment & \multicolumn{3}{c||}{24.0} & \multicolumn{3}{c}{73.5}\\\hline
\end{tabular}
\caption{Agreement of the weak labels, BERT-predicted labels, and FRESH-predicted labels with \textsc{NewsAgendas}' annotated gold labels. Metric reported is mean IOU/Recall-1 (standard deviation in parentheses for predicted labels).}
\label{tab:agreement}
\end{table*}

\subsection{Are the extracted rationales useful?}
We know that the FRESH rationales are useful to the BERT-predictors because our FRESH results show that BERT is able to achieve comparable prediction accuracy when using just the rationales as input as compared to using the entire text as input. Evaluating whether the FRESH rationales are also useful to humans is trickier. We analyze the percent of non-stopword rationale tokens that were also contained in the human-annotated rationales. However, we saw that the scores were not reliably different from just selecting the first 350 characters of the article as the rationale. This is likely because the generated rationales contain non-contiguous tokens from throughout the article, whereas the human-annotated rationales are 1-3 sentences. We therefore need to explore further human evaluation methods to quantitatively determine how well the model is rationalizing. 

Through manual inspection, the rationales also seem meaningful to a human. We show three examples of common scenarios in Table \ref{tab:rationales} that demonstrate the quality of the rationales and the low word overlap score with the human-annotated rationales. The first example in this table illustrates a case where the human and FRESH model chose different labels for the article but both labels and rationales seem reasonable. The second example shows a case where the human and FRESH model agreed on the label, and the model rationale actually shares almost all the major keywords of the human rationale (although these words are not contiguous and in the same order as in the case of the human rationale). The final example then shows a case where the human and FRESH model agreed on the label, but chose rationales with very few overlapping words other than \textit{Washington D.C.} and \textit{socialism}.
\begin{table*}[hbt!]
    \centering
    \begin{tabularx}{\textwidth}{p{4.2cm}|p{5.2cm}|p{5.3cm}}
    \hline
         \textbf{ Human-annotated} & \textbf{ Model-predicted} & \textbf{Article Opening} \\\hline\hline
          \small \textbf{Negative Sentiment}: American and global audiences have been bombarded with media images of wailing children in holding facilities, having been separated from adults (maybe their parents, maybe not) detained for illegal entry into the United States. & \small \textbf{Propaganda}: Atrocity Porn and Hitler Memes and Daddy ! since parents - caging children racist FDRs Indeed , voted for Trump is now Americans Nazis Hitler President Donald Trump First Lady ? sanctuary - border Trumps terrorism - His & \small \textbf{\hl{Atrocity Porn and Hitler Memes}} - American and global audiences have been bombarded with media images of wailing children in holding facilities, having been separated from adults (maybe their parents, maybe not) detained for illegal entry into the United States. The images have been accompanied by gut-wrenching audio...
          \\\hline
         \small \textbf{Junk Science}: Apple cider vinegar has so many benefits, but personally one of the reasons I like it best is because of the digestive and metabolism boosting benefits. & \small \textbf{Junk Science}: 6 Things Your Body Shot Apple Cider Vinegar Daily amazing health benefits apple cider vinegar ( acv salad health benefits apple cider vinegar Why ? well 1 active Metabolism It This body protein It Boosts helps iron oxygen Iron oxygen the oxygen muscle 5 It helps glucose 6 acids produce sleep a video...
         & \small \textbf{\hl{6 Things} That Happen To \hl{Your Body} When You Take A \hl{Shot} Of \hl{Apple Cider Vinegar Daily}} - We've all heard about the \hl{amazing health benefits} of \hl{apple cider vinegar (acv)} right? but we forget to take it or how to use it. Most people add a little bit to the top of their \hl{salad} when they remember but that's about it...\\\hline
          \small \textbf{Propaganda}: It isnnt any of their usual villains, the corporations and banks, who made Washington D.C. so miserable. It's the triumph of socialism. & \small \textbf{Propaganda}: Millionaire Poverty Pimps Fight 'Income Inequality' Washington D.C. citys food Obama Imperial City Income Inequality in America : Oligarchy Middle Class , the town hall oligarchic city income inequality government I nt Washington Its socialism bedroom left A socialist socialist socialist nt politics millionaire poverty & \small \textbf{\hl{Millionaire Poverty Pimps Fight 'Income Inequality'}} - The bottom fifth of \hl{Washington D.C.} account for just 2\% of the city's income. It has one of the highest poverty rates in the country and the highest \hl{food} stamp use. And under \hl{Obama}, the \hl{Imperial City} of the politicians and the poor was surrounded by some of the wealthiest districts in the country...\\\hline
    \end{tabularx}
    \caption{Examples of labels/rationales annotated by humans and predicted by FRESH. The FRESH rationale is a concatenation of the most salient words in the text, whereas the human-annotated rationale is typically a sentence. We also highlight the FRESH-rationale words in the article opening (the title and first couple sentences) for clarity.} 
    \label{tab:rationales}
\end{table*}

\section{Related Work}

\noindent
\textbf{Disinformation and Misinformation.} There are many previous approaches which have studied detection of misinformation and disinformation and which would be useful in combination with the detectors developed in this work (e.g., an agenda detection system flags an article to then go through a fact-checking pipeline). Research on detecting fake news includes detectors based on linguistic features \cite{gravanis2019behind}, fact-checking based systems \cite{ciampaglia2015computational}, social context or propagation network based approaches \cite{shu2020leveraging, wu2015false, liu2018early}, multi-modal approaches \cite{khattar2019mvae}, etc. Other work has focused on characterizing/defining disinformation as a whole and developing classification schemas for campaigns \cite{dichotomies, fallis2015disinformation}. However, neither disinformation detection nor characterization has explicitly looked at the more specific identification of a harmful agenda in an article. 

\noindent
\textbf{Intent Detection.} An agenda requires intention so detecting a harmful agenda is a type of intent detection. Intent detection is used in many settings with systems using slot-filling \cite{niu2019novel}, conversational techniques \cite{larson2019evaluation, casanueva2020efficient}, and language understanding \cite{qin2019stack}. There has also been research into what intentions are involved with news articles specifically - on the intention of writing vs. sharing articles \cite{yaqub2020effects}, the journalistic role of articles \cite{mellado2015professional, tsang2020issue}, and what motivates people to create and share fake news knowingly \cite{osmundsen2020partisan}. Finally, there has also been work on detecting deception (an intentional act) \cite{rubin2012discerning}. However, these works have not looked specifically at automatic classification of a harmful agenda in news.

\section{Conclusion}
In this work, we formalize the open challenge of detecting harmful agendas in news articles, release an initial evaluation dataset, and develop an interpretable system for this task. We hope our work can encourage future investment in this area - such as exploring state-of-the-art intepretable models for detecting the features we discussed, further characterizing article agenda beyond a binary classification, or investigating the interplay between text features and metadata like article source.


\section{Limitations}
Given the subjective nature of our proposed task, this work does have some limitations and challenges. Firstly, the notion of harm or potential to do harm is seldom an objective factor and is also difficult to measure or quantify. Our experiments on inter-annotator agreement use a small dataset, so this study could be expanded with collaboration with social science researchers to better qualify how people perceive the agenda in different articles. Our work is also grounded in the United States, so it may have limited applications to the news in other countries (discussed more in Section 8). Secondly, our data and framework can be used to build and train a system to perform post-hoc detection of harmful agendas in news articles. However, in a real-world system, this identification would likely need to happen on the fly, so as to make readers aware of these agendas as they are exposed to the articles. Finally, another aspect that we have not addressed in this study is the effect that a platform or community may have on the perceived harm in an article. For example, on dedicated social media channels hosting discussions on alternate theories and contentious topics (such as the efficacy of COVID-19 vaccines), a junk science article with dubious claims may not be as ``harmful" as opposed to the same article being posted on an open forum where readers may perceive it as scientific fact, thereby making the article more ``harmful". The context in which news articles are disseminated may have a profound impact on this perceived harm and this may be an interesting direction for future exploration.

\section{Ethical Considerations}
\label{eth}
\subsection{Censorship}
Detecting harmful agendas in news articles has the obvious possible downstream use of filtering or banning articles which are flagged as such from being shared on social media platforms. We have already seen debate over content filtering like this take place in relation to sites like Facebook, Instagram, and Twitter moderating the dissemination of ``fake news'' on their platforms. One could imagine an automatic harmful agenda detector becoming part of this kind of content moderation pipeline. However, if the AI system incorrectly flags articles, it may end up censoring legitimate political speech. For this reason, we discourage any real-world use of this system at this time until further research and analysis can be completed. Additionally, we want to emphasize that this detection system should be paired with a fact-checking system to make sure that the pipeline considers the interplay between agenda and misinformation, and does not just flag biased or opinionated free speech.

\subsection{Cultural/Ideological Context}
Characterizing an article as containing a harmful agenda forces definitions of what constitutes harm, which has been studied for millennia by philosophers of ethics. Normative ethics is the study of how to articulate the basic tenets of what is good and bad \citep{kagan2018normative}. Broadly, normative ethics is divided into teleological/consequentialist (focusing on consequences to determine good/bad \citep{sep-consequentialism}) and Deontological (moral worth is intrinsic to an action \citep{sep-ethics-deontological}). In this work, we focus on real-world harm which draws more on consequentialism. 

Ultimately, as these opposing theories demonstrate, there is no universal interpretation of good and bad, or scale for evaluating harm. For this reason, any attempt to characterize news articles will come from a certain cultural context and perspective. The dataset we present is subject to the biases and cultural contexts of the annotators involved, so while it represents a useful starting point for work and data collection efforts in this area, future datasets around this problem must be conscious of recruiting a diverse and large annotator pool. An example of an individual bias could be that for a devout believer in the Christian God, writing which denounces God's existence could be considered harmful disinformation. Whereas from the broader societal perspective of the United States, such a piece of writing would likely be considered a benign opinion piece.

Additionally, we want to clearly state that the framing of this research (in terms of what constitutes harm, fact, etc.) was through a United States sociopolitical context, and therefore likely does not apply across other global contexts without modifications. In conclusion, any future applications of news agenda characterization in the real world need to be very clear about the particular cultural context it is designed to operate in, what assumptions it uses, and what applications it is appropriate for.

\section*{Acknowledgments}

This work is supported by the DARPA SemaFor project (HR001120C0123). The authors would also like to thank journalism subject matter experts Kristy Roschke, Dan Gillmor and Scott Ruston for their insightful feedback and discussions. The views, opinions and/or findings expressed are those of the authors and should not be interpreted as representing the official views or policies of the Department of Defense or the U.S. Government.
\bibliography{anthology,custom}
\bibliographystyle{acl_natbib}

\appendix

\section{Training Hyperparameters}
We use BERT-for-Sequence-Classification (bert-base-cased) from Huggingface\footnote{\url{https://huggingface.co/docs/transformers/index}} for both the rationale extractor and the predictor, training on binary classification of the feature in question. We did not notice much sensitivity to hyperparameters during an initial grid-search, so we decided to use the AdamW optimizer with a learning rate of 1e-5; we applied an early stopping patience of 15 epochs and set the max number of epochs to be 50. All results are reported as an average with standard deviation across 3 different training runs (with random seeds 1000, 2000, 3000). We trained each FRESH model for several hours on 1 NVIDIA Titan Xp GPU. We also use the BERT models from the rationale extractor framework as a reference in our results since they are trained to predict the feature label from the article text. These BERT models are an artifact of training the FRESH models so they did not require additional computation.

\section{Annotator Recruitment and Training}
We posted a recruitment notice on a journalism ListServ. We then hired the first four students who responded who met the criteria of current students at the same university as one of the authors and native English speakers. We hired the students through the university and compensated them at a rate of \$20/hour for 9-12 hours of work each. This rate is above the minimum wage in the city where the students completed the work.

After completing hiring paperwork, students had a 1-on-1 call with one of the authors who explained the goal of the research and what the task would look like, and provided a chance to discuss concerns and questions. Throughout the process students could communicate with the authors at anytime over email with questions/concerns, and they could also opt-out of the work at anytime. Otherwise students were able to complete the work independently on the their own computers using the Amazon Mechanical Turk Workers Sandbox \footnote{https://workersandbox.mturk.com}. Students were compensated outside of the platform based on their hours, and no other workers on the platform completed the tasks.

\section{Annotator Instructions}
\begin{figure*}
\centering
    \includegraphics[width=0.9\textwidth]{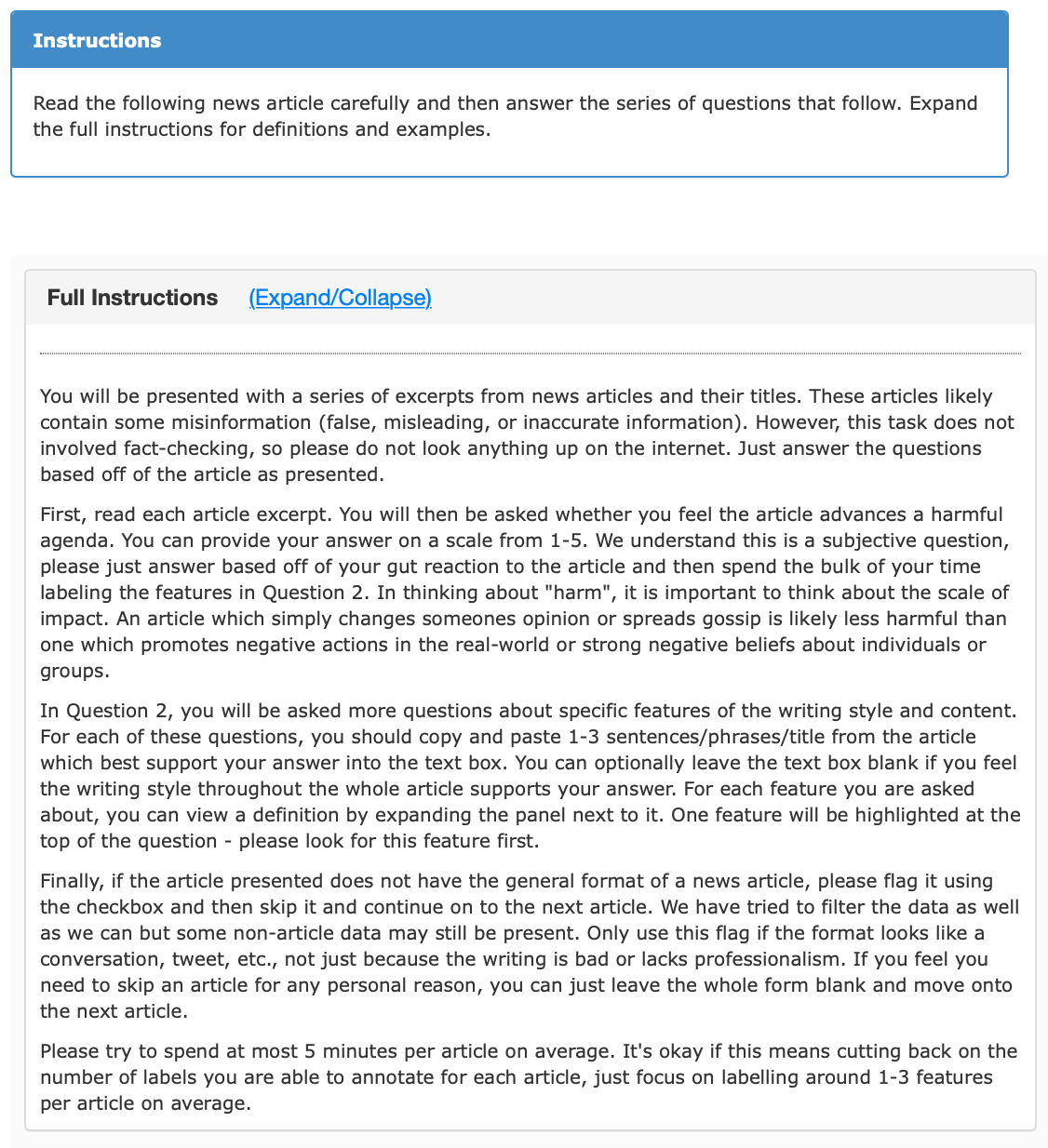}
\caption{The task instructions presented to annotators.}
\label{fig:instructions}
\end{figure*}

For the annotation of \textsc{NewsAgendas}, students were presented with the instructions shown in Figure \ref{fig:instructions}. They were not required to answer any of the questions, which allowed them to skip a whole article if the content made them uncomfortable since many of the articles contained offensive language.

\begin{figure*}
\centering
    \includegraphics[width=0.75\textwidth]{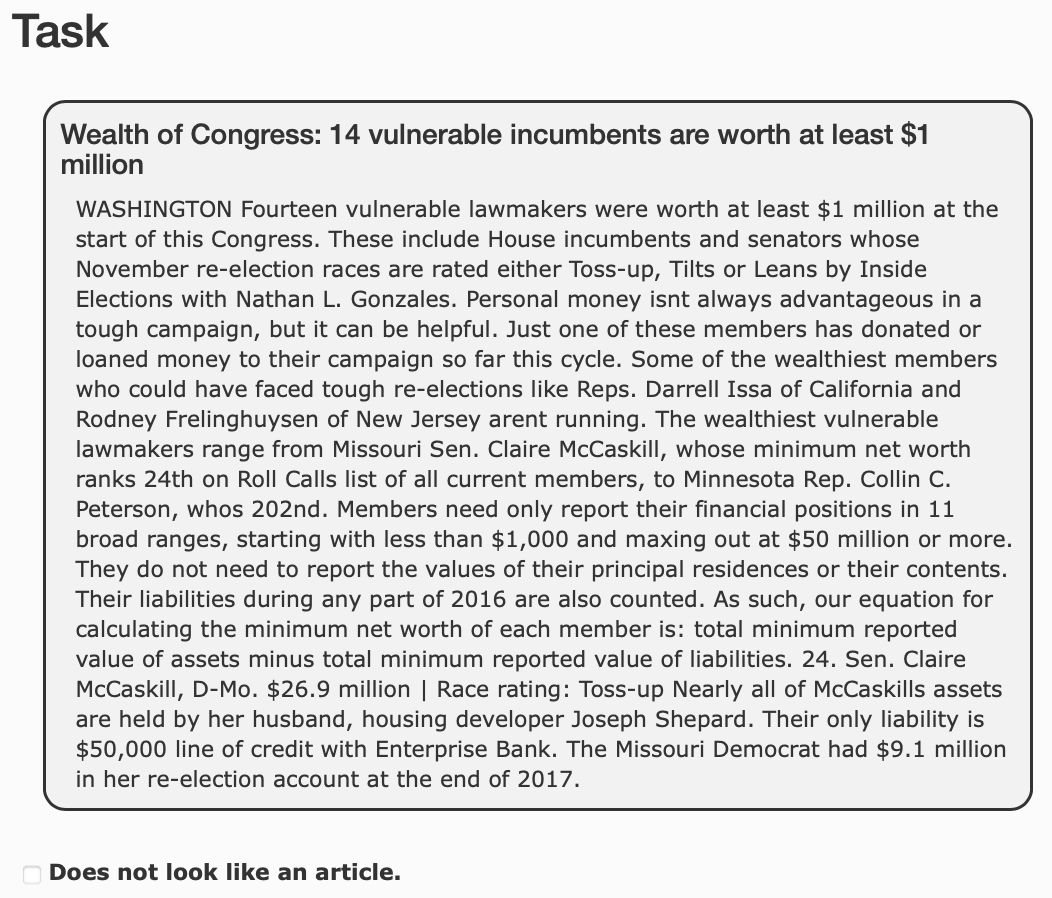}
\caption{An example article as shown in the task format.}
\label{fig:task}
\end{figure*}

\begin{figure*}
\centering
    \includegraphics[width=0.75\textwidth]{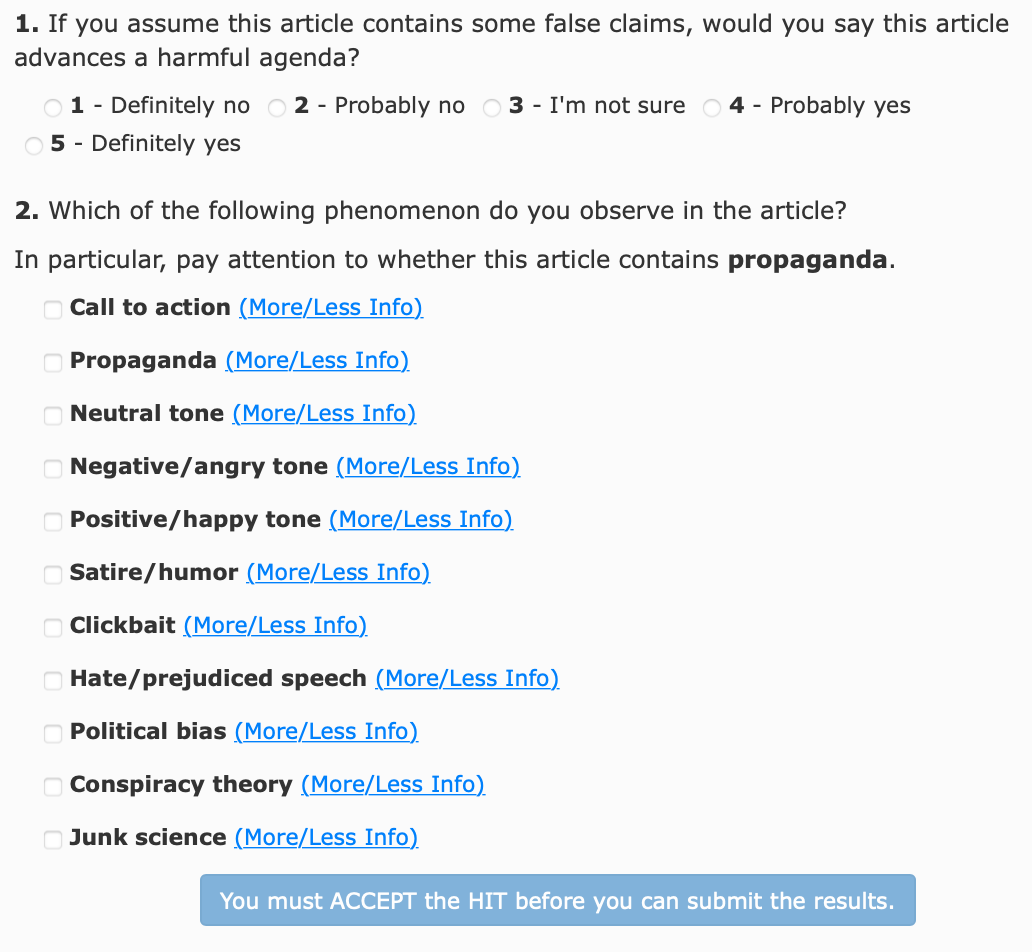}
\caption{The questions asked of annotators with an example weak label.}
\label{fig:questions}
\end{figure*}

\begin{figure*}
\centering
    \includegraphics[width=0.8\textwidth]{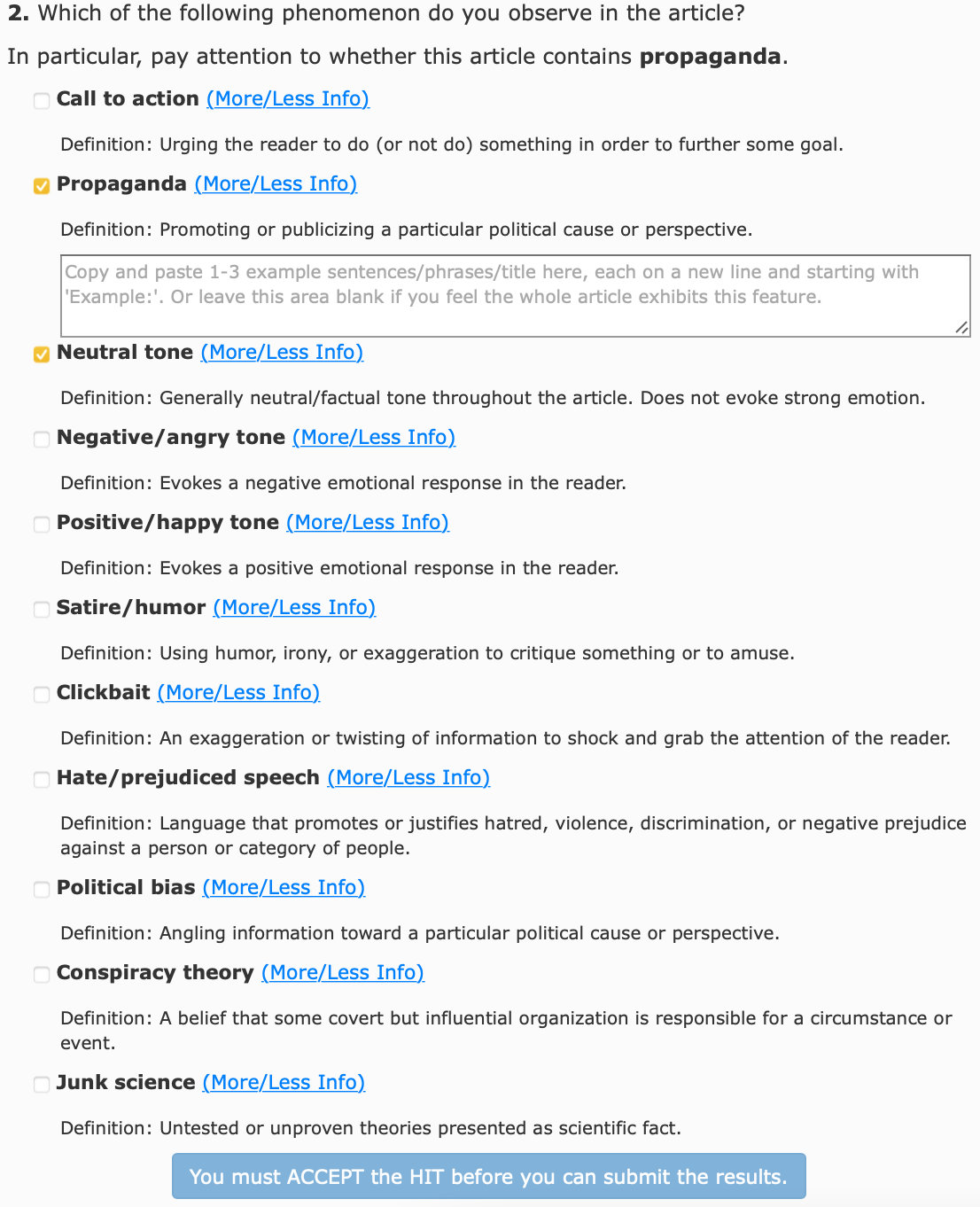}
\caption{An example of how the components in Question 2 could be expanded. Note that for `Neutral Tone', there was no option to provide evidence as this feature was generally present throughout the article. Otherwise, if an annotator selected a checkbox, the option to provide evidence would appear.}
\label{fig:expansion}
\end{figure*}

Articles were displayed to the annotators as shown in Figure \ref{fig:task}. They were then asked the questions shown in Figure \ref{fig:questions}. The feature names we used with the annotators differed slightly from the wording presented in this paper to facilitate clarity for the annotators. Whereas for this paper, we wanted to use consistent terminology throughout. Annotators could expand the label definitions in Question 2 as shown in Figure \ref{fig:expansion}.

We did not ask annotators any personal or demographic questions, and neither did we collect nor store any personal information about them.

\begin{figure*}
\centering
    \includegraphics[width=\textwidth]{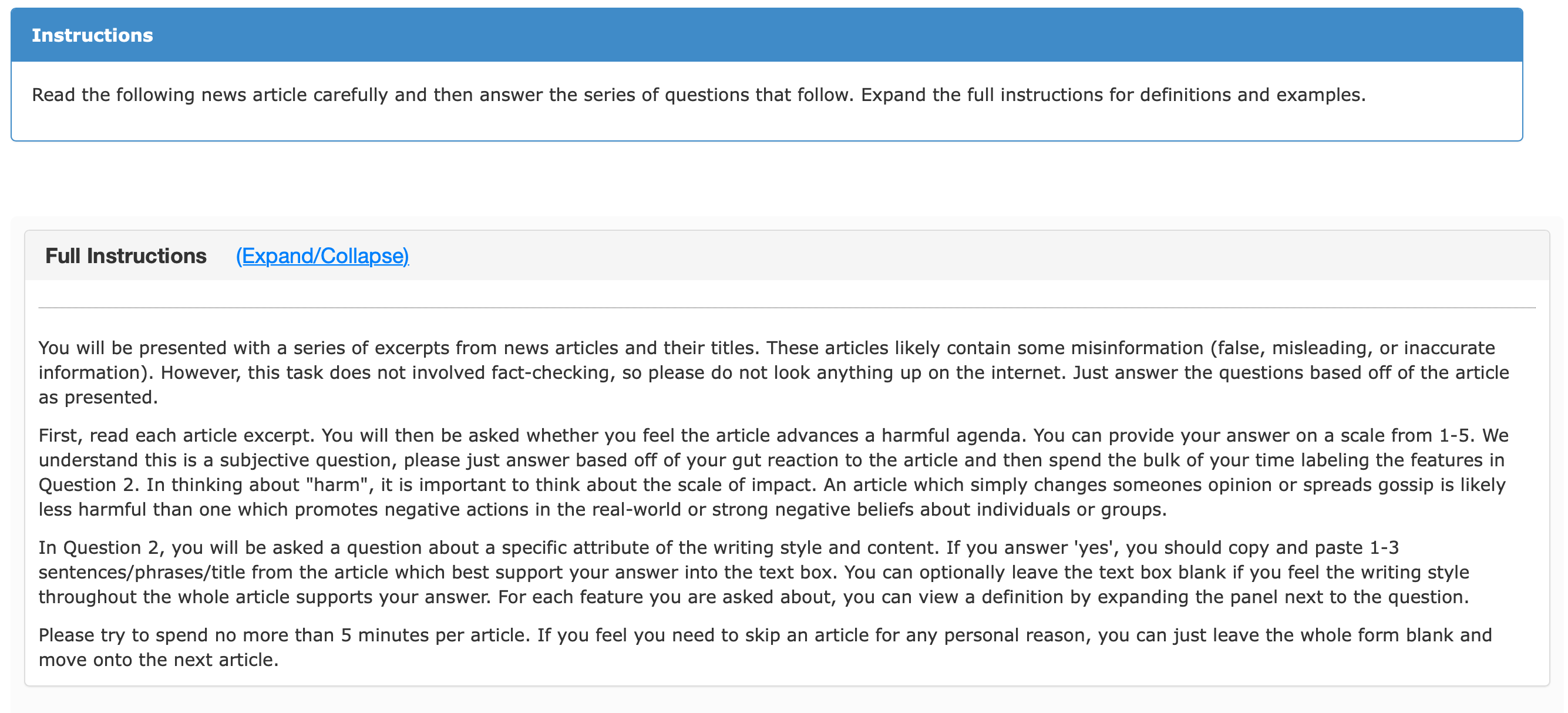}
\caption{The annotation quality experiment instructions.}
\label{fig:instructions2}
\end{figure*}

\begin{figure*}
\centering
    \includegraphics[width=\textwidth]{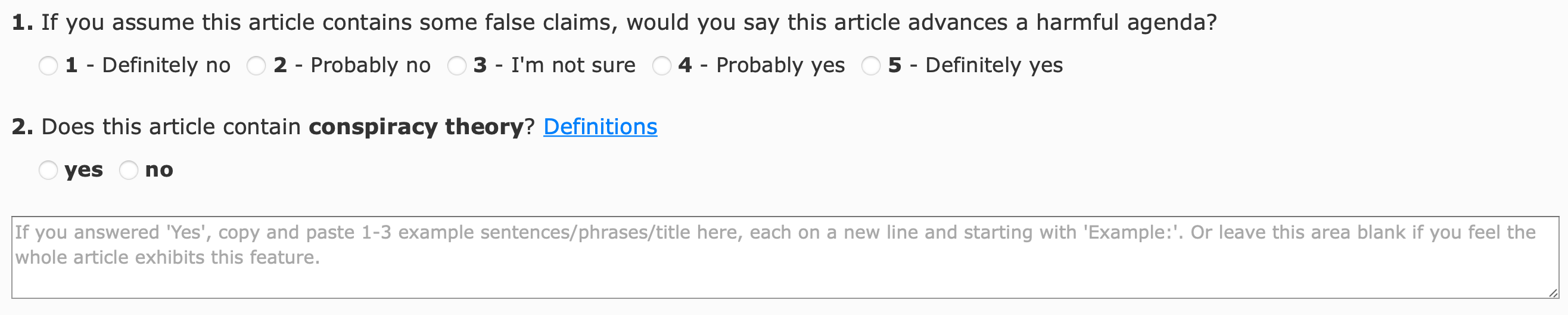}
\caption{The questions asked in the annotation quality experiments for most feature labels.}
\label{fig:attributeqs}
\end{figure*}

\begin{figure*}
\centering
    \includegraphics[width=\textwidth]{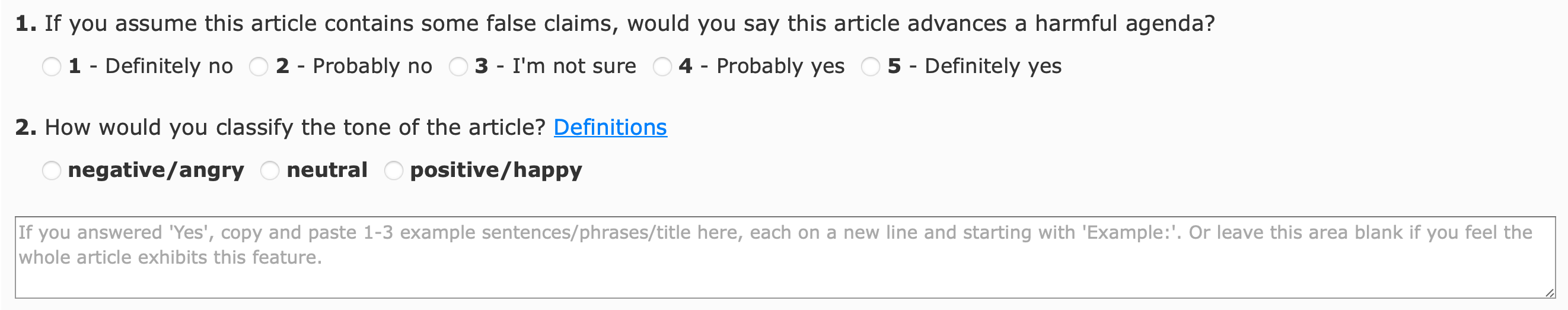}
\caption{The questions asked in the annotation quality experiments for tone-related labels.}
\label{fig:toneqs}
\end{figure*}

For the annotation quality experiments, students were presented with the instructions shown in Figure \ref{fig:instructions2}. Articles were displayed as shown in Figure \ref{fig:task}. The students were then asked the questions shown in Figure \ref{fig:attributeqs} for most feature labels, but the questions shown in Figure \ref{fig:toneqs} for tone-related labels. They could once again expand the definitions of the labels if needed.

\begin{figure*}
\centering
\begin{subfigure}[b]{0.47\textwidth}
\includegraphics[width=\textwidth]{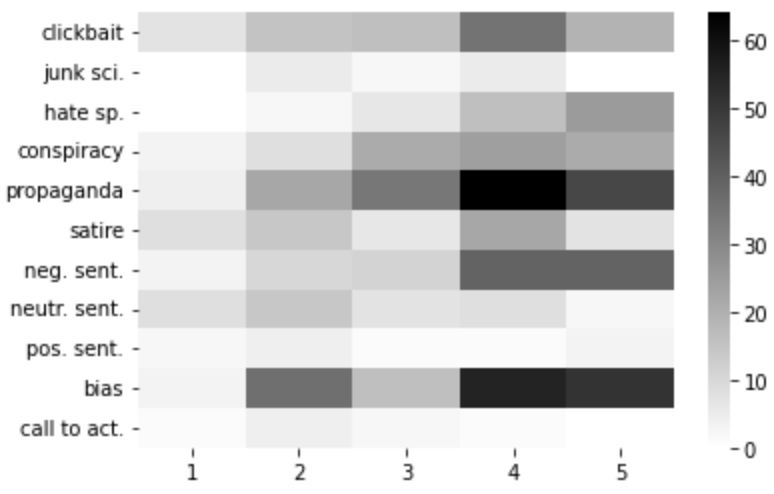}
\caption{Counts of each harmful agenda score associated with each feature label.}
\end{subfigure}
\hfill
\begin{subfigure}[b]{0.47\textwidth}
\includegraphics[width=\textwidth]{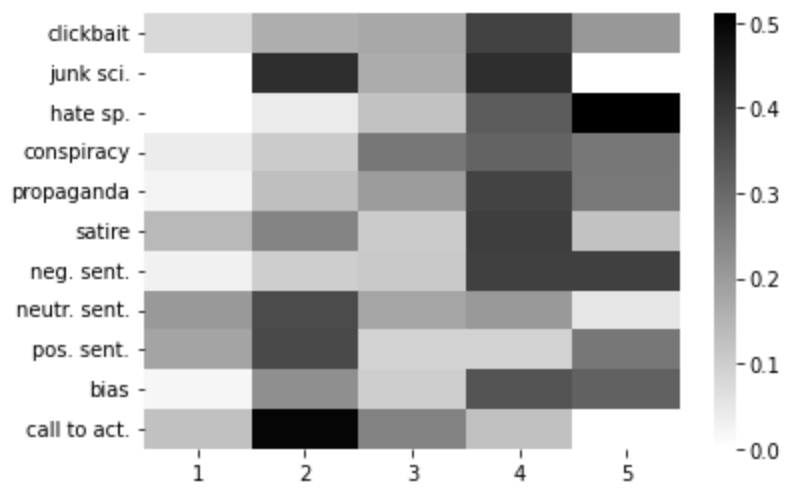}
\caption{Fraction of the agenda scores associated with each feature label that fall into each bucket. Each row sums to 1.}
\end{subfigure}
\caption{The distribution of agenda scores associated with each feature label.}
\label{fig:heatmaps}
\end{figure*}

\section{\textsc{NewsAgendas} Label Distribution and Examples}
The distribution of agenda scores labeled in \textsc{NewsAgendas} is shown in Table \ref{tab:datstats1}. The distribution of weak labels, annotated gold labels, and evidence spans for each feature is shown in Table \ref{tab:datstats2}. We also looked at the distribution of agenda scores across each feature, which is shown as heatmaps in Figure \ref{fig:heatmaps}. Examples of annotated evidence spans for each feature label are shown in Table \ref{tab:spanexamples}. 

\begin{table}[ht]
\centering
\begin{tabular}{c|c}
     \hline Agenda Score & Annotation Count  \\\hline\hline
     1 & 58 \\\hline 
     2 & 124 \\\hline
     3 & 87 \\\hline
     4 & 123 \\\hline
     5 & 69 \\\hline
\end{tabular}
\caption{Counts of the agenda scores labeled in \textsc{NewsAgendas}. There are also 45 data points for which annotators chose not to label an agenda score but selected feature labels and evidence spans.}
\label{tab:datstats1}
\end{table}

\begin{table}[ht]
\centering
\begin{tabularx}{0.5\textwidth}{l|c|c|c}
     \hline Feature&Weak& Annot. &  Spans \\\hline\hline
     Clickbait & 83&110&158 \\\hline
     Junk Science & 13&15&19 \\\hline
     Hate Speech & 4&54&65 \\\hline
     Conspiracy Theory & 52&84&102 \\\hline
     Propaganda & 220&198&289 \\\hline
     Satire & 52 & 64 & 104\\\hline
     Negative Sentiment & --&113&103 \\\hline
     Neutral Sentiment & --&42&-- \\\hline
     Positive Sentiment & --&13&14 \\\hline
    Political Bias & 35&181&234 \\\hline 
     Call to Action & --&8&10 \\\hline
\end{tabularx}
\caption{Counts of weak labels, annotated gold labels, and evidence spans for each feature in \textsc{NewsAgendas}.}
\label{tab:datstats2}
\end{table}

\begin{table*}[hbt!]
    \centering
    \begin{tabularx}{\textwidth}{p{3cm}|p{12.2cm}}
    \hline
         \textbf{ Label} & \textbf{ Example Spans from \textsc{NewsAgendas}}  \\\hline\hline
          Clickbait & \small \it Could \#RussianHackers have used a cloaking device to hide Wisconsin from Hillary? \\\hline
         Junk Science & \small \it Apple cider vinegar has so many benefits, but personally one of the reasons I like it best is because of the digestive and metabolism boosting benefits.  \\\hline
          Hate Speech & \small \it They are a race of ugly dwarves, of diminutive stature, with hideous faces, evil beady eyes and stunted small minds. \\\hline
         Conspiracy Theory & \small \it The case sparked national debate over immigration reform and so-called Sanctuary Cities that shield illegals from deportation, of which San Francisco is one. \\\hline
         Propaganda & \small \it President Barack Obama made sure to shutter veterans parks in an effort to make the GOP look bad during the shutdown which occurred under his watch.  \\\hline
         Satire & \small \it The former U.S. senator and former Democrat nominee for Vice President was charged with several felonies. Shockingly, felonious narcissism was not one of them. \\\hline
         Negative Sentiment & \small \it Once again, the party bereft of ideas and principle resorts to emotional obfuscation and accusation to advance their ideological prejudice. \\\hline
         Neutral Sentiment & \small \it A long lost Viking settlement known as `Hop' is located in Canada, a prominent archaeologist has revealed. \\\hline
         Positive Sentiment & \small \it Newspapers, pamphlets and broadsheets provided nourishment to both spark the American Revolution and keep it alive. \\\hline
         Political Bias & \small \it Although this news may sound surprising, there are valid reasons for blacks to gravitate toward Trump.  \\\hline
          Call to Action & \small \it We need your financial support to help reach those undecided voters, and if you would like to help, you can donate online right here. \\\hline
    \end{tabularx}
    \caption{Example evidence spans annotated in \textsc{NewsAgendas}.}
    \label{tab:spanexamples}
\end{table*}

\section{Negative Examples for Training Feature Detectors}
The challenge of negative sampling arises from the potential overlaps between the class labels. For example, an article can be both "junk science" and "conspiracy theory" in practice. In the FakeNewsCorpus, the websites (and thus the articles) can have multiple labels, including a primary label that best describes the source. However, these labels were based on annotators' overall impression of a website, which may not capture all possible types of its articles. Evidence suggests that websites sharing junk science articles often share conspiracy articles, or articles possessing both features (more details in the next paragraph). Then, even if a website has "junk science" as its only label, some of its articles may still be "conspiracy." Therefore, articles from this website may not be proper negative examples for a conspiracy detector.

With this observation, we develop our criteria for negative examples. For a model that detects a specific label (referred to as the positive label), we quantify the positive label's overlap with other class labels using the overlap coefficient (Szymkiewicz–Simpson coefficient). The overlap between Label A and Label B is calculated as $\frac{|A \cap B|}{\min(|A|,|B|) }$, where $A$ and $B$ are the sets of websites whose multiple labels include Label A and Label B respectively. After exploratory experiments on the validation set, we adopted a threshold of 0.15 to filter out classes that overlap too much with the positive class. For example, the overlap coefficient of "junk science" and "conspiracy" is 0.5396, exceeding 0.15. Thus, excluding "conspiracy" articles from the negative examples can better train the "junk science" detector. The negative classes after applying this criterion can be found in Table \ref{tab:negatives}. In addition to sampling from these selected negative classes, all negative samples must not have the positive label among their multiple labels. Since we have multiple negative classes, we include more negative examples than positive examples, depending on the availability of the former after applying the criteria. We adopt a standard class-weighted loss in training to handle class imbalance.

\begin{table*}[ht]
    \centering
    \begin{tabularx}{\textwidth}{p{2.9cm}|p{12.5cm}}
        \hline \textbf{Class} & \textbf{Negative Example Classes}  \\\hline\hline
        Clickbait & Conspiracy Theory, Hate Speech, Propaganda, Satire, Average\\\hline
        Junk Science & Hate Speech, Propaganda, Satire, Average\\\hline
        Hate Speech & Clickbait, Junk Science, Satire, Average \\\hline
        Conspiracy Theory & Clickbait, Satire, Average \\\hline
        Propaganda & Clickbait, Junk Science, Satire, Average\\\hline
        Satire & Clickbait, Conspiracy Theory, Hate Speech, Junk Science, Propaganda, Average\\\hline
    \end{tabularx}
    \caption{All training articles belong to one of the 7 classes - Clickbait, Junk Science, Hate Speech, Conspiracy, Propaganda, Satire - or are \textit{Average} articles, meaning likely truthful and informative. The class labels are from the FakeNewsCorpus, Proppy Corpus, and the \citet{yang_satirical_2017} satire dataset. We omit articles from websites that only have less informative labels such as \textit{bias} or \textit{political}.}
    \label{tab:negatives}
\end{table*}

\section{Additional Results and Analysis}
\begin{figure*}
\centering 
    \includegraphics[width=0.6\textwidth]{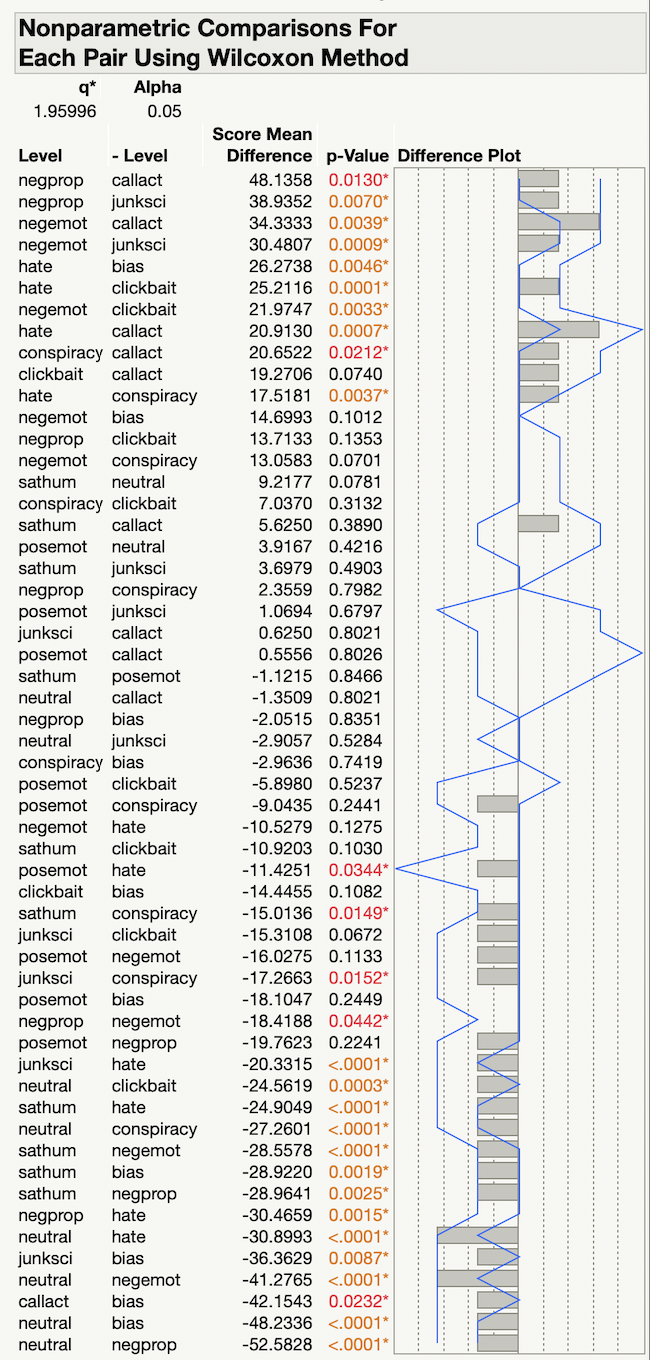}
\caption{Pairwise comparisons using the Wilcoxon method across the set of features with respect to the agenda score. A \textit{positive} Score Mean Difference with significant p-value implies that the articles with Label 1 are associated with higher agenda scores than articles with Label 2 (** $p<0.01$, * $p<0.05$). A \textit{negative} Score Mean Difference with significant p-values implies the opposite. The final column indicates the Score Mean Difference. The agenda score has a bi-modal distribution, as expected in Likert scale type survey responses. Key for feature names - negprop:propaganda, callact:call to action, negemot:negative sentiment, junksci:junk science, hate:hate speech, bias:political bias, clickbait:clickbait, conspiracy:conspiracy theories, neutral:neutral sentiment, sathum:satire, posemot:positive sentiment.} 
\label{fig:pairwise}
\end{figure*}

The full Wilcoxon pairwise comparisons (discussed in Section 4.2) are shown in Table \ref{fig:pairwise}.

\end{document}